\PassOptionsToPackage{table}{xcolor}
\documentclass[]{selfevolagent}

\usepackage{microtype}
\microtypesetup{expansion=false}
\usepackage{amsmath}
\usepackage{amssymb}
\usepackage{mathtools}
\usepackage[T1]{fontenc}
\usepackage[utf8]{inputenc}
\usepackage{array}
\usepackage{colortbl}
\usepackage{tabularx}
\usepackage{enumitem}
\usepackage{float}
\usepackage[percent]{overpic}
\usepackage{makecell}
\usepackage{balance}
\usepackage{fontawesome5}

\graphicspath{{figures/}}

\providecommand{\Description}[1]{}

\definecolor{tablegroup}{RGB}{244,247,251}
\definecolor{tablescaleFour}{RGB}{232,240,249}
\definecolor{tablescaleNine}{RGB}{232,244,239}
\definecolor{tablebest}{RGB}{218,239,226}
\definecolor{vadfigurenavy}{HTML}{14213D}
\definecolor{vadfiguregreen}{HTML}{08783E}
\definecolor{vadfigurepalegreen}{HTML}{F5FAF7}

\newcommand{\method}{VAD}
\newcommand{\methodlong}{Visual Attribution Distillation}
\newcommand{\bestscore}[1]{{\bfseries\boldmath #1}}
\newcommand{\secondscore}[1]{\underline{#1}}
\newcommand{\panelref}[2]{\hyperref[#1]{\ref*{#1}#2}}
\newcommand{\corrauth}{\text{\raisebox{-0.12ex}{\scalebox{0.78}{\faEnvelope}}}}

\floatstyle{ruled}
\newfloat{algorithm}{tbp}{loa}
\floatname{algorithm}{Algorithm}

\renewcommand\authorformat[2][]{{\fontsize{10}{13}\selectfont\sffamily\bfseries #2$^{#1}$}}

\title{VAD: Attributing Visual Evidence for Target Reconstruction in Multimodal On-Policy Distillation}
\author[1,2*]{Kangning Zhang}
\author[3]{~Yixing Li}
\author[1,2*]{~Shuai Shao}
\author[1,2*]{~Qingyao Li}
\author[4]{~Zhengxi Lu}
\author[4]{~Zhiyuan Yao}
\author[5]{~Shijian Wang}
\author[1]{~Jianghao Lin}
\author[2,\corrauth]{~Wenxiang Jiao}
\author[2,\corrauth]{~Yuan Lu}
\author[1,\corrauth]{~Weiwen Liu}
\author[1,\corrauth]{~Weinan Zhang}
\author[1,\corrauth]{~Yong Yu}

\affiliation[1]{Shanghai Jiao Tong University}
\affiliation[2]{Xiaohongshu Inc.}
\affiliation[3]{The Chinese University of Hong Kong}
\affiliation[4]{Zhejiang University}
\affiliation[5]{Southeast University}

\contribution[*]{Work done during internship at Xiaohongshu Inc.}
\contribution[\corrauth]{Corresponding authors}

\metadata[Contact]{zhangkangning@sjtu.edu.cn, wenxiangjiaonju@gmail.com, liuww@sjtu.edu.cn}
\metadata[Code]{\href{https://github.com/DeepExperience/VAD_Multimodal_OPD}{\nolinkurl{https://github.com/DeepExperience/VAD_Multimodal_OPD}}}
\metadata[Model]{\href{https://huggingface.co/zhangkangning/VAD_for_Qwen3.5-4b}{https://huggingface.co/zhangkangning/VAD\_for\_Qwen3.5-4b} \& \href{https://huggingface.co/zhangkangning/VAD_for_Qwen3.5-9b}{VAD\_for\_Qwen3.5-9b}}

\abstract{Multimodal on-policy distillation (OPD) transfers fine-grained visual knowledge by supervising student-generated trajectories with a privileged-view teacher. Yet its next-token corrections are \emph{source-mixed}, combining visual signals with linguistic priors and teacher-specific effects. The key challenge is to estimate which corrections are supported by visual evidence, not merely where or how strongly to distill.
We introduce \methodlong{} (\method{}), a counterfactual target-reconstruction algorithm that estimates the visually attributable part of a teacher correction. At each student-generated prefix, \method{} evaluates the same fixed teacher with the relevant evidence present and removed. The corresponding change in centered log-probabilities defines $u_t$, a signed proxy for the \emph{visual evidence direction} that estimates how revealing the evidence supports or refutes candidate tokens. \method{} projects the original correction onto this proxy to obtain an intervention-aligned component and a proxy-unexplained residual, then reconstructs a student-anchored target from the former. During training, this reconstructed target supplies the primary supervision signal, while the privileged teacher contributes a weak regularizer.
Across six fine-grained visual benchmarks at 4B and 9B scales, \method{} outperforms direct privileged-view distillation and visual-advantage weighting. Token-level and controlled-target analyses show that the proxy-aligned component is enriched in task-relevant visual corrections and yields stronger target shifts, especially when evidence refutes a mistaken answer. These results support counterfactual target reconstruction as an effective alternative to source-mixed supervision.}

\begin{document}

\maketitle

\section{Introduction}
\label{sec:introduction}

Many failures of multimodal large language models (MLLMs) begin with a small perceptual miss. A model may overlook a word, confuse an attribute, or misread a spatial relation; the remainder of its response can remain fluent while being grounded in the wrong visual evidence. Knowledge distillation transfers predictive structure from a stronger teacher \citep{hinton2015distillingknowledgeneuralnetwork,kim-rush-2016-sequence}, and on-policy distillation (OPD) makes this supervision relevant to deployment by querying the teacher on prefixes generated by the student itself \citep{agarwal2024onpolicydistillationlanguagemodels}. This property is especially valuable for multimodal reasoning, where an early perceptual error can redirect the entire generation trajectory.

\begin{figure*}[t]
    \centering
    \begin{overpic}[width=\textwidth]{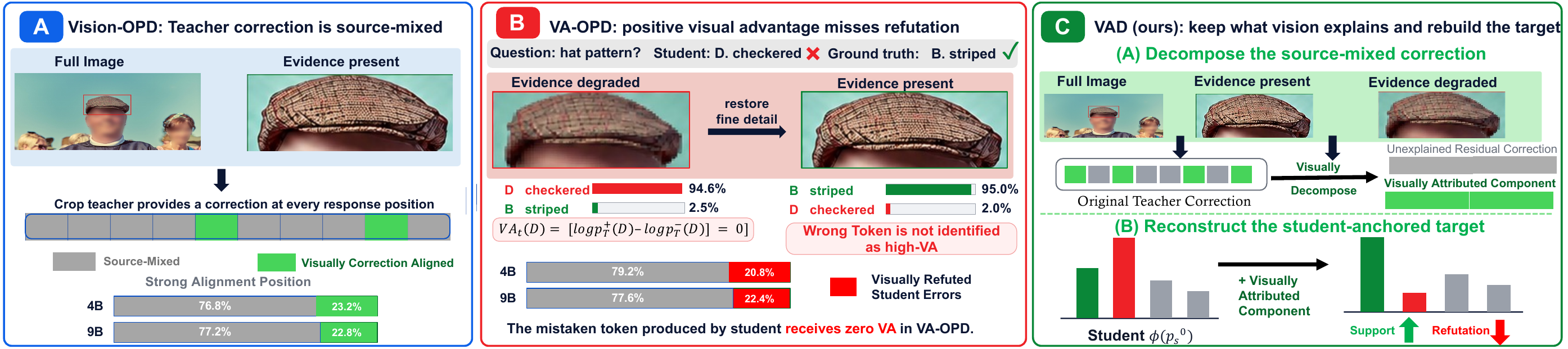}
      \setlength{\fboxsep}{0.5pt}
    \end{overpic}
    \Description{
    A three-panel overview showing source-mixed teacher correction in Vision-OPD,
    the refutation blind spot of positive visual advantage, and the correction
    decomposition and student-anchored target reconstruction used by VAD.
    }
    \caption{
    \textbf{Motivation and overview.}
    \textbf{(A)} A privileged-view teacher proposes a correction at every student-generated position, but strong teacher--student disagreement does not imply that the correction is strongly explained by visual evidence.
    \textbf{(B)} Positive visual advantage can miss refutation: revealing the relevant evidence may suppress the student's mistaken token and redirect probability toward a correct alternative.
    \textbf{(C)} \method{} uses an intervention-derived proxy for the visual evidence direction to estimate the visually attributable correction and reconstruct a student-anchored target.
    }
    \label{fig:introduction-overview}
\end{figure*}

Recent multimodal OPD mainly follows two strategies, whose limitations are illustrated in Figure~\ref{fig:introduction-overview}. First, Vision-OPD~\citep{yuan2026visionopdlearningfinedetails} conditions the teacher on an evidence-centered crop but directly distills its complete next-token distribution. In our diagnostic,\footnote{Appendix~\ref{sec:appendix-visual-alignment} details the diagnostic protocol and clustered uncertainty analysis.} only $23.2\%$ of the strongest response-token corrections for the 4B model and $22.8\%$ for the 9B model are strongly aligned with the teacher's evidence-conditioned response; the remaining high-correction positions are operationally source-mixed (Figure~\panelref{fig:introduction-overview}{A}). Second, VA-OPD~\citep{liu2026visualadvantageonpolicydistillationvisionlanguage} and V-Zero~\citep{sun2026vzeroanswerlabelfreeonpolicydistillation} contrast informative and degraded views to prioritize tokens or trajectories while retaining the whole evidence-present teacher as the underlying target. This improves selectivity but leaves source-mixed directions inside the target. Moreover, the positive visual advantage used by VA-OPD can miss refutation when clear evidence should lower the probability of the student's mistaken token (Figure~\panelref{fig:introduction-overview}{B}). These limitations expose a \emph{visual-correction attribution} problem: \textbf{estimating which component of the teacher's proposed correction is supported by a controlled visual intervention}.

We introduce \methodlong{} (\method{}) to estimate this attribution and use it to reconstruct the learning target. For each student-generated prefix, \method{} queries the same fixed teacher for the next-token distribution under an evidence-present view and an evidence-removed view while keeping the textual context unchanged. We denote the resulting shift in centered log-probabilities by $u_t$ and use it as an intervention-derived proxy for the \emph{visual evidence direction}. The proxy estimates how the chosen evidence intervention changes candidate-token odds; it is not an oracle grounding label or an exhaustive causal decomposition of visual information. It is inherently signed because revealing the evidence can support some candidate tokens while refuting others.

Guided by this proxy, \method{} applies a regularized projection to estimate the part of the original teacher correction aligned with the intervention; the remainder is treated as a proxy-unexplained residual, not as a purely nonvisual component. Rather than matching the complete privileged-teacher distribution, it reconstructs a target around the full-image student's current distribution using the proxy-aligned component. The target raises candidates that the intervention indicates are supported, suppresses those it indicates are refuted, and excludes correction components not captured by the proxy. \textbf{\method{} reconstructs what to distill rather than merely rescaling how strongly to distill}. To stabilize optimization, \method{} further retains a weak regularizer derived from the original teacher correction and splits $u_t$ into support and refutation branches, enabling separate control of candidates. The additional teacher views are required only during training; inference uses the standard full-image student.

We evaluate \method{} on six fine-grained visual benchmarks with 4B and 9B students. Across both scales, it consistently improves over direct privileged-view distillation and visual-advantage weighting. Candidate-token semantic analyses show that the intervention-derived proxy and its aligned correction are enriched in visual, task-relevant, and correct-answer tokens. Controlled offline studies further show that the reconstructed targets preserve useful visual correction.

Our contributions are threefold:
\begin{itemize}
    \item We identify \emph{source-mixed teacher correction} as a central limitation of privileged multimodal OPD and show why direct teacher matching and positive visual-advantage weighting do not explicitly estimate which correction component is supported by a controlled visual intervention.

    \item We introduce \method{}, which uses a controlled visual-evidence intervention to construct a proxy direction, estimates the aligned component of the privileged teacher correction, reconstructs a student-anchored target, separates support from refutation, and uses a weak teacher-correction regularizer for stable optimization.

    \item We demonstrate consistent gains across six benchmarks and two model scales, together with semantic token analyses and controlled target studies that support the intended interpretation of the proxy-aligned correction and its support--refutation behavior.
\end{itemize}

\section{Related Work}

\subsection{On-Policy and Privileged Distillation}

Knowledge distillation transfers soft teacher distributions~\citep{hinton2015distillingknowledgeneuralnetwork}, and sequence-level distillation extends it to autoregressive generation~\citep{kim-rush-2016-sequence}. Recent methods increasingly train on student-induced states: MiniLLM~\citep{gu2024minillm} optimizes reverse KL on student samples, GKD~\citep{agarwal2024onpolicydistillationlanguagemodels} unifies divergences and teacher--student data mixtures, DistiLLM~\citep{ko2024distillm} combines skew KL with adaptive student generations, and DistiLLM-2~\citep{ko2025distillm2} introduces a contrastive objective across teacher- and student-generated data. Their primary focus is the sampled trajectory, divergence, or data-source mixture.

Privileged distillation instead strengthens the teacher with information unavailable to the deployed student. On-Policy Context Distillation~\citep{ye2026onpolicycontext} internalizes solution histories or optimized prompts, Visual-OPSD~\citep{li2026visualopsd} transfers privileged visual-thought traces, Anchored Residual OPD~\citep{zhang2026absoluteimitationanchoredresidual} separates fully and partially privileged guidance, and Contrastive OPD~\citep{ruan2026contrastiveopd} contrasts light- and heavy-reasoning instructions. \method{} asks a complementary question: which part of a privileged teacher correction is attributable to a controlled visual-evidence change? Unlike Anchored Residual OPD's future-conditioned textual residual, \method{} compares evidence-present and evidence-removed views under the same teacher and prefix, then reconstructs a signed visual target around the full-image student.

\subsection{Fine-Grained Visual Perception and Privileged Evidence}

Instruction-tuned MLLMs support open-ended visual reasoning~\citep{liu2023visualinstructiontuning,dai2023instructblipgeneralpurposevisionlanguagemodels,bai2023qwenvlversatilevisionlanguagemodel}, while Qwen2-VL~\citep{wang2024qwen2vlenhancingvisionlanguagemodels}, Qwen2.5-VL~\citep{bai2025qwen25vltechnicalreport}, LLaVA-UHD~\citep{xu2024llavauhd}, and Mini-Gemini~\citep{li2024minigemini} improve native resolution or multi-scale detail. Nevertheless, diagnostic and high-resolution benchmarks continue to reveal failures on decisive local evidence~\citep{tong2024eyeswideshutexploring,chen2024rightwayevaluatinglarge,wang2024divideconquercombinetrainingfree,zhang2025mmerealworldmultimodalllmchallenge}.

One strategy actively acquires better views: V*~\citep{wu2023vguidedvisualsearch} and ZoomEye~\citep{shen-etal-2025-zoomeye} search image regions, ReFocus~\citep{fu2025refocus} edits images as visual thoughts, and DeepEyes~\citep{zheng2026deepeyes}, Thyme~\citep{zhang2025thyme}, DeepEyesV2~\citep{hong2026deepeyesv2}, and SenseNova-MARS~\citep{chng2026sensenovamars} invoke image operations or multimodal tools. These methods expose fine detail at the cost of inference-time search or tool use.

Region-to-image training instead internalizes privileged local evidence. Zooming without Zooming~\citep{wei2026zoomingzoomingregiontoimagedistillation} distills region-centric perception into a full-image pass; Vision-OPD~\citep{yuan2026visionopdlearningfinedetails} uses a crop-conditioned self-teacher along student rollouts; and ViCuR~\citep{tian2026vicurvisualcuesrecoverable} trains the student to recover privileged visual cues. Whereas these works improve evidence acquisition or representation, \method{} addresses target construction by excluding teacher changes not explained by the revealed evidence.

\subsection{Counterfactual Visual Supervision and Multimodal OPD}

Visual interventions expose weak grounding through controlled evidence changes. Visual Contrastive Decoding~\citep{leng2024visualcontrastivedecoding} contrasts original and distorted images at inference, HALC~\citep{chen2024halc} adds focal visual contrast, and later analysis compares the signals induced by downsampling and image editing~\citep{lee2024delvevisualcontrastivedecoding}. Thinking with Deltas~\citep{gao2026thinkingdeltas} uses original, masked, and perturbed images during reinforcement learning. These methods target decoding or outcome-driven optimization rather than dense on-policy target construction.

Multimodal OPD uses visual contrast in several ways. Vision-OPD \citep{yuan2026visionopdlearningfinedetails} directly matches the evidence-present teacher. VA-OPD \citep{liu2026visualadvantageonpolicydistillationvisionlanguage} and Med-OPD~\citep{qian2026medopd} use evidence-aware rollout or token weighting. V-Zero \citep{sun2026vzeroanswerlabelfreeonpolicydistillation} forms a group-relative trajectory gate, while DOPD \citep{yu2026dopddualonpolicydistillation} routes token supervision between teacher and student policies. They determine \emph{where, how strongly, or from which policy} to distill, but do not reconstruct the privileged target from its visually attributable correction.

Decomposed OPD \citep{yoon2026decomposedonpolicydistillationvisionlanguage} factorizes multimodal distillation into language-prior and visual-grounding objectives. It constructs a visual target from the student's text-only prior and the teacher's multimodal-to-text-only information gain, then uses Visual Gradient Steering to combine this target with standard distillation. \method{} addresses a different attribution problem: it projects the privileged teacher-to-student correction onto a same-teacher evidence-present versus evidence-removed direction, anchors the reconstructed target at the full-image student, and explicitly budgets visual support and refutation. Thus, both methods reconstruct visual supervision, but Decomposed OPD prioritizes general visual-information matching whereas \method{} selects which part of a privileged correction is warranted by the controlled evidence intervention.

\section{Visual Attribution Distillation}
\label{sec:method}

\begin{figure*}[t]
  \centering
  \begin{overpic}[width=\textwidth]{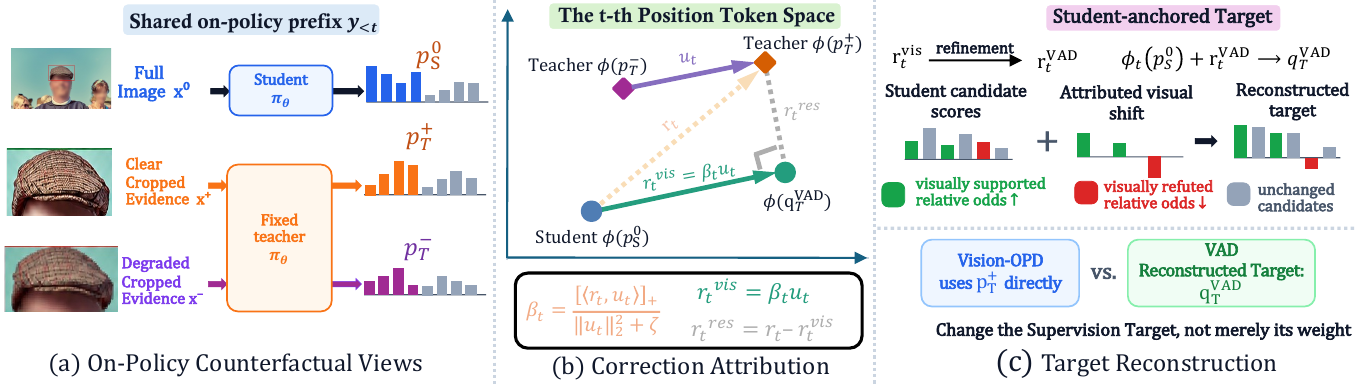}
    \setlength{\fboxsep}{0.5pt}
  \end{overpic}
  \Description{A three-stage overview of VAD. A student response prefix is reused to obtain aligned next-token distributions from the full-image student and a fixed teacher under clear and degraded visual views. The teacher correction is attributed to the counterfactual visual response, producing a signed visual shift. This shift is added to the student-centered logits to reconstruct a new distillation target, with a weak teacher-target stability anchor used during training.}
  \caption{\textbf{Overview of \method{}.} \textbf{(a) On-policy counterfactual views.} Given the same student-generated prefix, the full-image student produces its next-token distribution, while a fixed teacher evaluates evidence-present and evidence-removed views under the unchanged textual context. \textbf{(b) Correction attribution.} The teacher-view contrast defines an intervention-derived proxy $u_t$ for the visual evidence direction; projecting the complete teacher correction $r_t$ onto this proxy yields the attributed component $r_t^{\mathrm{vis}}$ and the proxy-unexplained residual $r_t^{\mathrm{res}}$. \textbf{(c) Target reconstruction.} \method{} refines $r_t^{\mathrm{vis}}$ into a budgeted signed shift that supports or refutes candidate tokens and adds it to the student-centered logits to construct $q_{T,t}^{\mathrm{VAD}}$. The auxiliary teacher views are used only during training.}
  \label{fig:method-overview}
\end{figure*}

\method{} learns from on-policy student prefixes while using counterfactual teacher views only to construct supervision. Figure~\ref{fig:method-overview} gives the conceptual pipeline, and Algorithm~\ref{alg:vad} states the complete training procedure, including its inputs, reconstructed target, optimization objective, and inference-time output. The following subsections define each operation.

\subsection{On-Policy Counterfactual Views}

Let $x^0$ denote the full image available to the student. The student policy $\pi_{\theta}$ samples a response $y\sim\pi_{\theta}(\cdot\mid x^0)$, and all teacher queries reuse the student prefix $y_{<t}$. We keep a fixed copy $\pi_{\bar\theta}$ of the initial model as the teacher and construct two training-only views: an evidence-present crop $x^+$ and an evidence-removed or degraded crop $x^-$. At position $t$, the three next-token distributions are
\begin{equation}
\begin{aligned}
p_S^0 &= \pi_{\theta}(\cdot\mid x^0,y_{<t}),\\
p_T^+ &= \pi_{\bar\theta}(\cdot\mid x^+,y_{<t}),\\
p_T^- &= \pi_{\bar\theta}(\cdot\mid x^-,y_{<t}).
\end{aligned}
\label{eq:three-distributions}
\end{equation}
Because the teacher parameters and text prefix are identical, the contrast $p_T^+-p_T^-$ isolates the teacher's distributional response when the relevant visual evidence is made visible.

The training implementation evaluates all three distributions on a compact shared coordinate set $V_t$ constructed from the student's top-$K$ candidates. We gather the teacher probabilities on the same token identifiers and map each restricted distribution to a centered log-probability vector in $\mathbb{R}^{|V_t|}$:
\begin{equation}
\phi_t(p)
 =
\log\left(p[V_t]+\epsilon\right)
-
\operatorname{mean}\left(\log\left(p[V_t]+\epsilon\right)\right).
\label{eq:centered-log-probability}
\end{equation}
Centering removes a common logit offset while preserving pairwise log-odds on the shared support.

\subsection{Attributing the Teacher Correction}

The privileged-teacher correction and its response to the visual intervention are
\begin{equation}
 r_t=\phi_t(p_T^+)-\phi_t(p_S^0),
 \qquad
 u_t=\phi_t(p_T^+)-\phi_t(p_T^-).
 \label{eq:teacher-and-visual-directions}
\end{equation}
Here $r_t$ is the complete correction prescribed by the privileged teacher. For a candidate token $i$, $u_t(i)>0$ means that revealing the evidence raises its relative probability, whereas $u_t(i)<0$ means that the evidence refutes it; $|u_t(i)|$ measures the strength of this response at prefix $y_{<t}$. \textbf{We therefore use $u_t$ as a proxy vector for the core visual information available at position $t$, rather than as an oracle grounding label or a pixel gradient}.

We retain the part of $r_t$ that agrees with this visual response through a one-sided projection:
\begin{equation}
 \beta_t
 =
 \frac{[\langle r_t,u_t\rangle]_+}
 {\lVert u_t\rVert_2^2+\zeta},
 \qquad
 r_t^{\mathrm{vis}}=\beta_tu_t,
 \qquad
 r_t^{\mathrm{res}}=r_t-r_t^{\mathrm{vis}},
 \label{eq:one-sided-attribution}
\end{equation}
where $[a]_+=\max(a,0)$ and $\zeta$ stabilizes the projection. If $r_t$ and $u_t$ do not agree, $\beta_t=0$ and no visual shift is applied. Otherwise, the same nonnegative coefficient raises visually supported token odds and lowers visually refuted token odds. This gives the one-sided supervision target
\begin{equation}
 q_{T,t}^{\mathrm{one}}
 =
 \operatorname{softmax}\!\left(
 \phi_t(p_S^0)
 +
 \operatorname{clip}(r_t^{\mathrm{vis}},-c,c)
 \right).
 \label{eq:one-sided-target}
\end{equation}
Unlike direct teacher matching, this target starts from the student's current distribution and changes only the token odds attributed to the visual intervention. The residual $r_t^{\mathrm{res}}$ denotes the remaining, unattributed correction and is not assumed to be purely linguistic. We retain $B_t=\lVert r_t^{\mathrm{vis}}\rVert_2$ as the visual-correction budget for the refinement below.

\subsection{Budgeted Support and Refutation}

A lightweight refinement decouples the support and refutation coordinates that share one coefficient in the one-sided target. We define
\begin{equation}
 \begin{gathered}
 u_t^+=[u_t]_+,
 \qquad
 u_t^-=[u_t]_-,
 \qquad
 s_t^{\pm}=[\langle r_t,u_t^{\pm}\rangle]_+,\\
 Z_t=s_t^++s_t^-+\epsilon,
 \qquad
 \omega_t^+=\min\!\left(\frac{s_t^+}{Z_t},\tau_+\right),
 \qquad
 \omega_t^-=\frac{s_t^-}{Z_t},
 \end{gathered}
 \label{eq:budgeted-branch-weights}
\end{equation}
where $[u_t]_+=\max(u_t,0)$ and $[u_t]_- = \min(u_t,0)$. The agreement scores allocate $B_t$ between the two signed branches. We cap only the support share to prevent uncertain positive evidence from dominating; removed mass is not reassigned. The resulting correction and target are
\begin{equation}
 \begin{aligned}
 r_t^{\mathrm{VAD}}
 &=
 B_t\left(
 \omega_t^+\frac{u_t^+}{\lVert u_t^+\rVert_2+\epsilon}
 +
 \omega_t^-\frac{u_t^-}{\lVert u_t^-\rVert_2+\epsilon}
 \right),\\
 q_{T,t}^{\mathrm{VAD}}
 &=
 \operatorname{softmax}\!\left(
 \phi_t(p_S^0)
 +
 \operatorname{clip}(r_t^{\mathrm{VAD}},-c,c)
 \right).
 \end{aligned}
 \label{eq:vad-target}
\end{equation}
This refinement changes how the one-sided visual budget is divided, not the underlying attribution principle or the student-anchored target construction.
Appendix~\ref{sec:appendix-token-level-comparison} gives the corresponding token-wise odds identities and distinguishes target reconstruction from the rollout and token-group aggregation used by VA-OPD.

\subsection{Training Objective and Inference}

Given reconstructed target $q$ and student distribution $p$, let $m=(q+p)/2$. Our primary supervision is the standard token-level Jensen--Shannon objective used in OPD,
\begin{equation}
 D_{\mathrm{JS}}(q,p)
 =
 \frac{1}{2}\left[
 D_{\mathrm{KL}}(q\Vert m)
 +
 D_{\mathrm{KL}}(p\Vert m)
 \right].
 \label{eq:js-divergence}
\end{equation}
\begin{equation}
 \mathcal{L}_{\mathrm{vis}}
 =
 \frac{1}{|\mathcal{T}|}
 \sum_{t\in\mathcal{T}}
 D_{\mathrm{JS}}\!\left(
 \operatorname{stopgrad}(q_{T,t}^{\mathrm{VAD}}),
 p_S^0
 \right),
 \label{eq:primary-vad-objective}
\end{equation}
where $\mathcal{T}$ is the set of valid response positions. Relative to standard OPD, the divergence is unchanged; only the target is replaced by $q_{T,t}^{\mathrm{VAD}}$.

\textbf{Using $\mathcal{L}_{\mathrm{vis}}$ alone concentrates supervision on $r_t^{\mathrm{vis}}$}. In our visual-only ablation, the policy consequently exhibited clear language and output drift: responses became substantially longer and more repetitive, answer commitment was delayed, and formatting and stopping became unstable. This behavior is consistent with the construction of the target. Visual attribution preserves the evidence-sensitive correction but does not directly restore language realization, answer format, or EOS behavior outside that component.

We therefore add a weak privileged-teacher regularizer in formal training. Its token weight is larger when a smaller fraction of the complete correction is visually attributed:
\begin{equation}
\begin{aligned}
 \rho_t
 &=
 \frac{\lVert r_t^{\mathrm{VAD}}\rVert_2}
 {\lVert r_t\rVert_2+\epsilon},
 \qquad
 a_t
 =
 \operatorname{stopgrad}\!\left(
 \operatorname{clip}(1-\rho_t,0,1)
 \right),\\
 \mathcal{L}_{\mathrm{reg}}
 &=
 \frac{1}{|\mathcal{T}|}
 \sum_{t\in\mathcal{T}}
 a_t
 D_{\mathrm{JS}}\!\left(
 \operatorname{stopgrad}(p_T^+),
 p_S^0
 \right),
 \qquad
 \mathcal{L}
 =
 \mathcal{L}_{\mathrm{vis}}
 +
 \lambda\mathcal{L}_{\mathrm{reg}}.
\end{aligned}
\label{eq:regularized-vad-objective}
\end{equation}
The regularizer weakly anchors semantics, formatting, response length, and stopping while leaving the reconstructed visual target as the primary signal. All target-side quantities, including the copy of $p_S^0$ used to construct $q_{T,t}^{\mathrm{VAD}}$, are detached during optimization. Numerical settings are reported in Section~\ref{sec:experimental-setup}.

The teacher, crop, and degraded view are used only to construct training targets. At inference time, \method{} is a standard full-image student policy and introduces no additional model call or visual view.

\subsection{End-to-End Training Procedure}

Algorithm~\ref{alg:vad} summarizes the complete training loop. The reconstructed distribution supplies the primary JSD supervision, while the privileged teacher contributes only the weak regularizer.

\begin{algorithm}[H]
  \caption{Training \method{} with counterfactual target reconstruction}
  \label{alg:vad}
  \normalsize
  \renewcommand{\arraystretch}{1.08}
  \begin{tabularx}{\columnwidth}{@{}r@{\hspace{0.45em}}X@{}}
    & \textbf{Input:} training set $\mathcal{D}$; student $\pi_{\theta}$; fixed teacher $\pi_{\bar\theta}$; view constructor $\mathcal{V}(x^0)=(x^+,x^-)$; target and regularization parameters. \\
    & \textbf{Output:} optimized full-image student $\pi_{\theta}$; no teacher or auxiliary view is used at inference. \\
    1 & \textbf{for each} minibatch $\mathcal{B}\subset\mathcal{D}$ \textbf{do} \\
    2 & \quad \textbf{for each} full-image input $x^0\in\mathcal{B}$ \textbf{do} \\
    3 & \qquad Sample $y\sim\pi_{\theta}(\cdot\mid x^0)$ and construct $(x^+,x^-)=\mathcal{V}(x^0)$. \\
    4 & \qquad \textbf{for} $t=1,\ldots,|y|$ \textbf{do} \\
    5 & \qquad\quad Evaluate $(p_S^0,p_T^+,p_T^-)$ on $y_{<t}$ and compute their centered log probabilities on $V_t$. \\
    6 & \qquad\quad Form $r_t$ and $u_t$ by Eq.~\eqref{eq:teacher-and-visual-directions}, obtain $r_t^{\mathrm{vis}}$ by Eq.~\eqref{eq:one-sided-attribution}, and set $B_t=\lVert r_t^{\mathrm{vis}}\rVert_2$. \\
    7 & \qquad\quad Allocate the support and refutation budget by Eq.~\eqref{eq:budgeted-branch-weights} and construct $q_{T,t}^{\mathrm{VAD}}$ by Eq.~\eqref{eq:vad-target}. \\
    8 & \qquad\quad Accumulate the primary JSD and weak regularizer in Eq.~\eqref{eq:regularized-vad-objective}. \\
    9 & \qquad \textbf{end for} \\
    10 & \quad \textbf{end for} \\
    11 & \quad Update $\theta$ with the minibatch loss and keep $\bar\theta$ fixed. \\
    12 & \textbf{end for} \\
  \end{tabularx}
\end{algorithm}

\section{Experiments}

\begin{table*}[!t]
  \centering
  \normalsize
  \setlength{\tabcolsep}{2.5pt}
  \renewcommand{\arraystretch}{1.12}
  \caption{Overall comparison on six fine-grained visual benchmarks. Rows above the Qwen3.5-based block provide cross-family capability context; rows within each Qwen3.5 scale use matched data and post-training budgets. All entries use the same official evaluation pipeline. Values are accuracies (\%), and $\operatorname{Avg}_6$ is the unweighted mean. Within each Qwen3.5 scale, best results are bold and second-best results are underlined; cyan-green rows denote \method{}.}
  \label{tab:main-results}
  \begin{tabularx}{\textwidth}{@{}lc*{7}{>{\centering\arraybackslash}X}@{}}
    \toprule
    {\footnotesize Model} & {\footnotesize Param.} & {\footnotesize V$^\star$} & {\footnotesize Zoom} & \shortstack{\footnotesize HR-Bench\\[-1pt]\footnotesize 4K} & \shortstack{\footnotesize HR-Bench\\[-1pt]\footnotesize 8K} & \shortstack{\footnotesize MME-RW\\[-1pt]\footnotesize EN} & \shortstack{\footnotesize MME-RW\\[-1pt]\footnotesize CN} & {\footnotesize $\operatorname{Avg}_6$} \\
    \midrule
    \rowcolor{tablegroup}
    \multicolumn{9}{c}{\itshape ``Thinking-with-Images'' Agentic Models} \\
    DeepEyes~\citep{zheng2026deepeyes} & 7B & $82.63$ & $46.15$ & $75.50$ & $70.00$ & $63.95$ & $62.41$ & $66.77$ \\
    Thyme-RL~\citep{zhang2025thyme} & 7B & $78.47$ & $46.23$ & $78.87$ & $70.87$ & $64.15$ & $61.41$ & $66.67$ \\
    DeepEyesV2~\citep{hong2026deepeyesv2} & 7B & $77.95$ & $46.11$ & $79.75$ & $72.62$ & $64.25$ & $61.89$ & $67.09$ \\
    SenseNova-MARS~\citep{chng2026sensenovamars} & 8B & $88.42$ & $48.95$ & $85.00$ & $77.25$ & $67.25$ & $65.72$ & $72.10$ \\
    \midrule
    \rowcolor{tablegroup}
    \multicolumn{9}{c}{\itshape Closed-Source Models} \\
    Gemini 3 Flash~\citep{googledeepmind2025gemini3flash} & -- & $85.39$ & $61.49$ & $87.25$ & $85.00$ & $73.15$ & $71.66$ & $77.32$ \\
    Gemini 3.1 Pro~\citep{googledeepmind2026gemini31pro} & -- & $88.48$ & $62.01$ & $88.12$ & $84.50$ & $73.32$ & $71.81$ & $78.04$ \\
    GPT-5.2~\citep{openai2025gpt52} & -- & $84.59$ & $54.71$ & $84.64$ & $78.48$ & $71.82$ & $67.43$ & $73.61$ \\
    GPT-5.4~\citep{openai2026gpt54} & -- & $85.83$ & $57.94$ & $86.15$ & $78.10$ & $74.16$ & $70.50$ & $75.45$ \\
    \midrule
    \rowcolor{tablegroup}
    \multicolumn{9}{c}{\itshape Large Open-Source Models} \\
    Qwen3-VL-Instruct~\citep{bai2025qwen3vltechnicalreport} & 235B & $87.37$ & $57.23$ & $88.00$ & $79.25$ & $71.09$ & $65.86$ & $74.80$ \\
    Qwen3.5-397B-A17B~\citep{qwenteam2026qwen35} & 397B & $84.23$ & $58.30$ & $91.25$ & $84.37$ & $74.17$ & $66.64$ & $76.49$ \\
    GLM-4.6V~\citep{zhipuai2026glm46v} & 106B & $83.20$ & $52.59$ & $83.87$ & $80.13$ & $66.23$ & $66.30$ & $72.07$ \\
    Kimi-K2.6~\citep{moonshotai2026kimik26} & 1T & $83.77$ & $54.67$ & $83.62$ & $79.25$ & $69.88$ & $66.81$ & $73.00$ \\
    \midrule
    \rowcolor{tablegroup}
    \multicolumn{9}{c}{\itshape Qwen3.5-Based Models} \\
    \rowcolor{tablegroup}
    \multicolumn{9}{c}{\bfseries 4B Scale} \\
    Qwen3.5~\citep{qwenteam2026qwen35} & 4B & $82.20$ & $48.28$ & \bestscore{$85.75$} & \secondscore{$81.38$} & $63.89$ & $63.58$ & $70.85$ \\
    GRPO~\citep{shao2024deepseekmathpushinglimitsmathematical} & 4B & $81.68$ & $56.57$ & $76.75$ & $75.62$ & $71.10$ & $68.58$ & $71.72$ \\
    VA-OPD~\citep{liu2026visualadvantageonpolicydistillationvisionlanguage} & 4B & \secondscore{$90.05$} & $58.11$ & $81.88$ & $78.75$ & $73.29$ & $67.42$ & $74.92$ \\
    V-Zero~\citep{sun2026vzeroanswerlabelfreeonpolicydistillation} & 4B & $88.48$ & $56.45$ & $84.88$ & $78.75$ & $73.49$ & \secondscore{$70.45$} & $75.42$ \\
    Vision-OPD~\citep{yuan2026visionopdlearningfinedetails} & 4B & $89.53$ & \secondscore{$59.41$} & $81.75$ & $80.12$ & \secondscore{$74.51$} & $70.17$ & \secondscore{$75.92$} \\
    Decomposed OPD~\citep{yoon2026decomposedonpolicydistillationvisionlanguage} & 4B & $89.80$ & $58.70$ & $82.60$ & $79.50$ & $73.80$ & $67.82$ & $75.37$ \\
    \rowcolor{tablebest}
    \textbf{\method{} (Ours)} & 4B & \bestscore{$92.15$} & \bestscore{$60.59$} & \secondscore{$85.38$} & \bestscore{$83.38$} & \bestscore{$76.97$} & \bestscore{$71.47$} & \bestscore{$78.32$} \\
    \cmidrule(lr){1-9}
    \rowcolor{tablegroup}
    \multicolumn{9}{c}{\bfseries 9B Scale} \\
    Qwen3.5~\citep{qwenteam2026qwen35} & 9B & $86.39$ & $52.31$ & $85.12$ & $80.88$ & $71.13$ & $67.31$ & $73.86$ \\
    GRPO~\citep{shao2024deepseekmathpushinglimitsmathematical} & 9B & $89.10$ & $57.41$ & $85.12$ & $82.75$ & $73.48$ & $69.35$ & $76.20$ \\
    VA-OPD~\citep{liu2026visualadvantageonpolicydistillationvisionlanguage} & 9B & $86.91$ & \bestscore{$62.84$} & \secondscore{$86.12$} & $82.75$ & $71.17$ & $70.20$ & $76.67$ \\
    V-Zero~\citep{sun2026vzeroanswerlabelfreeonpolicydistillation} & 9B & $89.53$ & $57.04$ & $85.50$ & \secondscore{$83.88$} & \secondscore{$75.29$} & \secondscore{$71.52$} & \secondscore{$77.13$} \\
    Vision-OPD~\citep{yuan2026visionopdlearningfinedetails} & 9B & \secondscore{$92.72$} & $59.26$ & $85.13$ & $83.75$ & $71.76$ & $68.65$ & $76.88$ \\
    Decomposed OPD~\citep{yoon2026decomposedonpolicydistillationvisionlanguage} & 9B & $91.40$ & $60.20$ & $85.80$ & $83.10$ & $72.70$ & $69.10$ & $77.05$ \\
    \rowcolor{tablebest}
    \textbf{\method{} (Ours)} & 9B & \bestscore{$94.76$} & \secondscore{$62.49$} & \bestscore{$87.88$} & \bestscore{$85.75$} & \bestscore{$76.56$} & \bestscore{$72.13$} & \bestscore{$79.93$} \\
    \bottomrule
  \end{tabularx}
\end{table*}

\begin{figure*}[!t]
  \centering
  \includegraphics[width=\textwidth]{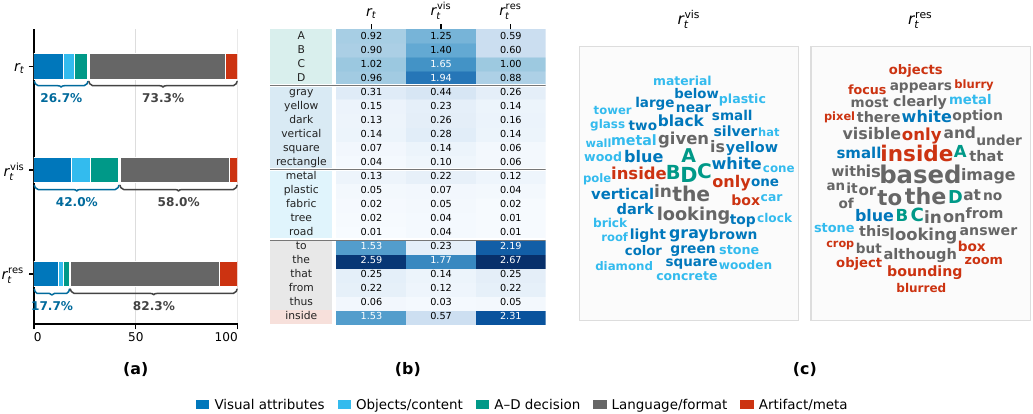}
  \Description{Three panels compare the complete teacher correction, its visually attributed component, and the residual. The visual component contains relatively more visual attributes, task objects, and answer-decision tokens, while the residual contains relatively more language, formatting, and teacher-specific tokens.}
  \caption{\textbf{Semantic separation of the teacher correction.} \textbf{(a)} Relative top-$5$ composition of the complete, visual, and residual directions after excluding Other. \textbf{(b)} Representative token-level component scores. \textbf{(c)} High-magnitude lexical profiles for $r_t^{\mathrm{vis}}$ and $r_t^{\mathrm{res}}$.}
  \label{fig:source-mixed-diagnosis}
\end{figure*}

We organize the experiments around five research questions:
\begin{itemize}
    \item[\textbf{RQ1.}] Does \method{} improve fine-grained visual accuracy over scale-matched post-training baselines?
    \item[\textbf{RQ2.}] Does the decomposition concentrate visual evidence in $r_t^{\mathrm{vis}}$ while shifting broader linguistic and teacher-specific semantics toward $r_t^{\mathrm{res}}$?
    \item[\textbf{RQ3.}] How do alternative supervision targets affect trained accuracy, correct-token support, and wrong-token suppression?
    \item[\textbf{RQ4.}] How do checkpoints produced by different post-training algorithms generalize to held-out tasks?
    \item[\textbf{RQ5.}] How do the supervision divergence, regularization weight $\lambda$, and positive-branch cap $\tau_+$ affect training outcomes?
\end{itemize}

\subsection{Setup}
\label{sec:experimental-setup}

\paragraph{Training setup.}
We train Qwen3.5-4B and Qwen3.5-9B on the same $6{,}241$ synthetic visual question-answering examples released by Vision-OPD~\citep{yuan2026visionopdlearningfinedetails}; no training image or question overlaps the six primary benchmarks. Each example provides a full image $x^0$ with the relevant region marked, an evidence-present $2\times$ crop $x^+$, and an evidence-degraded view $x^-$ obtained by $0.1\times$ bilinear downsampling and nearest-neighbor upsampling of that crop. The latter two views share region and size but differ in fine-grained evidence. A frozen copy of the initial student serves as teacher. Following Section~\ref{sec:method}, we optimize token-level JSD on reconstructed targets over the student top-$100$ support plus a tail bucket. We use batch size $96$, $8$ rollouts per prompt, learning rate $2\times10^{-6}$, projection stabilizer $\zeta=10^{-3}$, coordinate bound $c=20$, and weak-regularizer weight $\lambda=0.1$; the positive-branch cap $\tau_+$ is $0.8$ for 4B and $0.7$ for 9B.

Context rows include ``thinking with images'' agents \citep{zheng2026deepeyes,zhang2025thyme,hong2026deepeyesv2,chng2026sensenovamars}, closed-source Gemini models \citep{googledeepmind2025gemini3flash,googledeepmind2026gemini31pro}, and large open-source models \citep{bai2025qwen3vltechnicalreport,qwenteam2026qwen35,zhipuai2026glm46v,moonshotai2026kimik26}; differences in architecture, data, and compute preclude controlled comparison. Controlled rows instead share Qwen3.5 initialization \citep{qwenteam2026qwen35}, training examples, and post-training budget across GRPO \citep{shao2024deepseekmathpushinglimitsmathematical}, Vision-OPD \citep{yuan2026visionopdlearningfinedetails}, VA-OPD \citep{liu2026visualadvantageonpolicydistillationvisionlanguage}, V-Zero \citep{sun2026vzeroanswerlabelfreeonpolicydistillation}, Decomposed OPD \citep{yoon2026decomposedonpolicydistillationvisionlanguage}, and \method{}; We followed the official training script and further tuned the hyperparameters accordingly. Appendix~\ref{sec:appendix-controlled-algorithms} details their objectives and optimization roles. We evaluate VStar, ZoomBench \citep{wu2023vguidedvisualsearch,wei2026zoomingzoomingregiontoimagedistillation}, HRBench (4K/8K) \citep{wang2024divideconquercombinetrainingfree}, and MME-RealWorld (EN/CN) \citep{zhang2025mmerealworldmultimodalllmchallenge}. All results use the official Vision-OPD inference and accuracy pipeline with GPT-OSS-120B~\citep{openai2025gptossmodelcard} as judge; Table~\ref{tab:main-results} reports the six accuracies and their unweighted mean $\operatorname{Avg}_6$.

\subsection{RQ1: Fine-Grained Visual Accuracy}
\label{sec:rq1-main-results}

\textbf{With matched data and update budgets, \method{} is the strongest post-training method at both scales.}
It reaches $78.32$ and $79.93$ in $\operatorname{Avg}_6$, leading the best scale-matched alternative by $2.40$ points at 4B and $2.80$ points at 9B. Against the closely related Decomposed OPD baseline, the gains are $2.95$ and $2.88$ points, respectively. These controlled results isolate the benefit of attributing and reconstructing the privileged correction rather than using additional examples or updates. Appendix~\ref{sec:appendix-training-cost} further compares realized wall-clock and GPU cost for the available runs.

\textbf{The advantage persists against much larger open- and closed-source systems.}
Even the 4B checkpoint exceeds Gemini 3 Flash, Gemini 3.1 Pro, and Qwen3.5-397B, whose averages are $77.32$, $78.04$, and $76.49$; the 9B checkpoint raises the score to $79.93$. These cross-family comparisons demonstrate parameter efficiency, while the matched rows provide the controlled evidence for the training objective.

\textbf{The gains are broad rather than driven by a single benchmark.}
\method{} exceeds both Vision-OPD and Decomposed OPD on every benchmark at both scales. It leads every controlled post-training method on all six 4B evaluations; at 9B, it leads five of six, with only ZoomBench favoring VA-OPD by $0.35$ points. The improvements span fine-detail localization, high-resolution perception, and real-world recognition rather than concentrating on one dataset.

\subsection{RQ2: Semantics of the Attributed Correction}
\label{sec:rq2-correction-semantics}

Beyond downstream accuracy, we ask what semantic content the decomposition assigns to each supervision direction. At the same student-generated prefixes, we analyze the original privileged-teacher correction $r_t$, the component $r_t^{\mathrm{vis}}$ attributed to the counterfactual visual response, and the residual $r_t^{\mathrm{res}}=r_t-r_t^{\mathrm{vis}}$. Figure~\ref{fig:source-mixed-diagnosis} examines these directions at three complementary levels: category composition, representative token weights, and high-magnitude lexical profiles. This analysis tests whether attribution relocates visually grounded, decision-relevant content rather than merely shrinking the original correction.

\textbf{The attributed direction selectively concentrates visual evidence and decision semantics.}
In Figure~\panelref{fig:source-mixed-diagnosis}{(a)}, visual attributes, objects/content, and A--D decisions jointly account for $42.0\%$ of the relative top-$5$ composition in $r_t^{\mathrm{vis}}$, compared with $26.7\%$ in $r_t$ and only $17.7\%$ in $r_t^{\mathrm{res}}$. Correspondingly, language/format and artifact/meta semantics decrease to $58.0\%$ in $r_t^{\mathrm{vis}}$, versus $73.3\%$ in $r_t$ and $82.3\%$ in $r_t^{\mathrm{res}}$. This selective redistribution shows that the projection concentrates visually grounded, decision-relevant content rather than uniformly rescaling the teacher correction.

\textbf{The token-level evidence confirms a semantic separation rather than a scalar rescaling.}
Figure~\panelref{fig:source-mixed-diagnosis}{(b)} shows that every A--D answer symbol, together with representative attribute and object tokens such as \texttt{vertical} and \texttt{metal}, is more prominent in $r_t^{\mathrm{vis}}$; language scaffolding and artifact-related tokens such as \texttt{to} and \texttt{inside} are instead more prominent in $r_t^{\mathrm{res}}$. The broader lexical profiles in Figure~\panelref{fig:source-mixed-diagnosis}{(c)} exhibit the same organization: $r_t^{\mathrm{vis}}$ foregrounds colors, shapes, materials, spatial relations, objects, and answer choices, whereas $r_t^{\mathrm{res}}$ foregrounds linguistic realization and prompt- or degradation-related content. Taken together, the three views support the intended interpretation: $r_t^{\mathrm{vis}}$ carries the visually attributable evidence-to-decision signal, while $r_t^{\mathrm{res}}$ retains the portion of the teacher correction not explained by the visual intervention; the latter should therefore be interpreted as a mixed residual, not as purely linguistic noise.

\subsection{RQ3: Effects of Alternative Supervision Targets}
\label{sec:rq3-target-construction}

We compare six objectives with matched Qwen3.5-4B initialization, data, fixed teacher, and post-training budget:
\begin{itemize}[leftmargin=1.2em,labelsep=0.35em,itemsep=0pt,parsep=0pt,topsep=2pt]
  \item \emph{Direct}: uses the complete privileged-teacher distribution $p_T^+$.
  \item \emph{Scalar-shrunk}: rescales $r_t$ to its visually attributed norm but keeps the source-mixed direction.
  \item \emph{One-sided w/o regularization}: uses $q_T^{\mathrm{one}}=\allowbreak\operatorname{softmax}\!\bigl(\phi_t(p_S^0)+\allowbreak\operatorname{clip}(r_t^{\mathrm{vis}},-c,c)\bigr)$, where $r_t^{\mathrm{vis}}=\beta_tu_t$, without the teacher anchor.
  \item \emph{One-sided}: adds the weak teacher anchor to the same $q_T^{\mathrm{one}}$ target.
  \item \emph{\method{} w/o regularization / Full \method{}}: both use the branch-separated target $q_T^{\mathrm{VAD}}$; the latter adds the same anchor.
\end{itemize}
We organize the comparison around two questions: (1) under the same post-training budget, how do alternative supervision targets affect the accuracy of the trained model; and (2) at fixed answer-token positions in an offline analysis, how strongly does each target increase the probability of the correct answer and decrease that of the student's chosen wrong answer?

\begin{table}[!t]
  \centering
  \small
  \setlength{\tabcolsep}{0.2pt}
  \renewcommand{\arraystretch}{1.16}
  \caption{Target-construction and regularization ablation with Qwen3.5-4B under official scoring. Values are accuracies (\%). Best results are bold and second-best results are underlined; the cyan-green row denotes the full \method{} objective.}
  \label{tab:target-construction-ablation}
  \begin{tabularx}{\columnwidth}{@{}l*{7}{>{\centering\arraybackslash}X}@{}}
    \toprule
    Target & V$^\star$ & Zoom & \shortstack{HR-\\4K} & \shortstack{HR-\\8K} & \shortstack{MME-\\EN} & \shortstack{MME-\\CN} & $\operatorname{Avg}_6$ \\
    \midrule
    Direct $p_T^+$ & $89.53$ & $59.41$ & $81.75$ & $80.12$ & $74.51$ & $70.17$ & $75.92$ \\
    Scalar-shrunk & $90.61$ & $59.49$ & $82.24$ & $79.71$ & $74.67$ & $70.42$ & $76.19$ \\
    One-sided w/o reg. & \secondscore{$92.40$} & $59.58$ & $84.90$ & $79.95$ & $74.85$ & $70.70$ & $77.06$ \\
    One-sided & \bestscore{$93.19$} & $59.64$ & \secondscore{$86.12$} & $80.12$ & $75.07$ & \secondscore{$70.98$} & $77.52$ \\
    \method{} w/o reg. & $92.06$ & \secondscore{$60.57$} & \bestscore{$86.75$} & \secondscore{$82.74$} & \secondscore{$75.92$} & $70.29$ & \secondscore{$78.06$} \\
    \rowcolor{tablebest}
    \textbf{Full \method{}} & $92.15$ & \bestscore{$60.59$} & $85.38$ & \bestscore{$83.38$} & \bestscore{$76.97$} & \bestscore{$71.47$} & \bestscore{$78.32$} \\
    \bottomrule
  \end{tabularx}
\end{table}

\textbf{Branch-aware target reconstruction provides the main gain, with a smaller benefit from regularization.}
Table~\ref{tab:target-construction-ablation} reports official six-benchmark accuracy under matched training conditions. Scalar shrinking improves direct matching by only $0.27$ points in $\operatorname{Avg}_6$. Without regularization, replacing $q_T^{\mathrm{one}}$ with $q_T^{\mathrm{VAD}}$ raises the average from $77.06$ to $78.06$, isolating a $1.00$-point gain from branch-aware reconstruction. The weak anchor further raises the one-sided and \method{} targets to $77.52$ and $78.32$, respectively. Full \method{} therefore achieves the best average, $0.80$ points above regularized one-sided projection and $0.26$ points above its target-only counterpart.

\begin{figure*}[!t]
  \centering
  \begin{minipage}[t]{0.485\textwidth}
    \vspace{0pt}
    \centering
    \includegraphics[width=\linewidth]{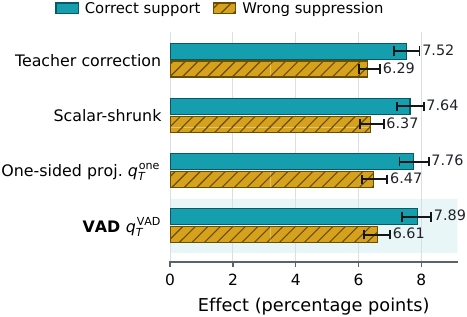}
    \Description{A horizontal grouped bar chart compares the offline probability gained by the correct answer token and removed from the student's chosen wrong token under the teacher correction, scalar shrinking, the one-sided target, and the branch-separated VAD target. Each bar has a 95 percent confidence interval, and the VAD target row is lightly highlighted.}
    \captionof{figure}{\textbf{Offline answer-token effects of supervision directions.} Bars show correct-token support and wrong-token suppression (percentage points); whiskers denote $95\%$ confidence intervals. The $q_T^{\mathrm{one}}$ and $q_T^{\mathrm{VAD}}$ rows exclude the weak teacher regularizer.}
    \label{fig:target-effects}
  \end{minipage}
  \hfill
  \begin{minipage}[t]{0.485\textwidth}
    \vspace{0pt}
    \centering
    \includegraphics[width=\linewidth]{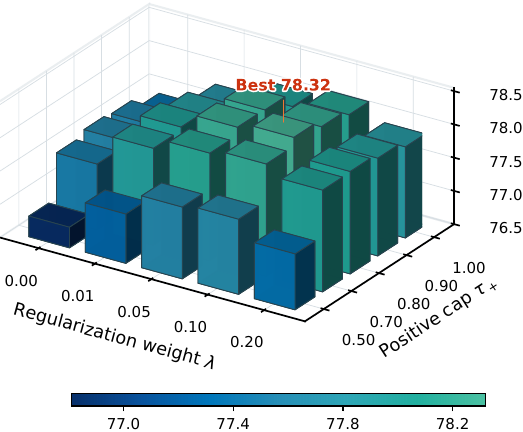}
    \Description{A five-by-five three-dimensional bar chart reports average accuracy for every combination of regularization weight and positive-branch cap. The best measured setting is regularization weight 0.10 and positive cap 0.80, with an average of 78.32.}
    \captionof{figure}{\textbf{Effect of $\lambda$ and $\tau_+$ on Qwen3.5-4B training.} Each bar reports $\operatorname{Avg}_6$ (\%) under the official evaluation pipeline.}
    \label{fig:parameter-sensitivity}
  \end{minipage}
\end{figure*}

\textbf{The reconstructed \method{} target yields the largest joint offline effect.}
Figure~\ref{fig:target-effects} isolates four target directions at fixed answer-slot states. Correct-token support increases monotonically from $7.52$ points for the teacher correction to $7.89$ for $q_T^{\mathrm{VAD}}$, while wrong-token suppression increases from $6.29$ to $6.61$ points; scalar shrinking and $q_T^{\mathrm{one}}$ lie between these endpoints. Because the confidence intervals overlap, the result supports a consistent directional trend rather than pairwise statistical superiority. Excluding the weak regularizer attributes this trend specifically to target construction.

\subsection{RQ4: Generalization of Post-Trained Checkpoints}
\label{sec:rq4-held-out-generalization}

We test whether the learned policies transfer beyond the fine-grained suite using MMVP~\citep{tong2024eyeswideshutexploring}, CV-Bench~\citep{tong2024cambrian1fullyopenvisioncentric}, MMStar~\citep{chen2024rightwayevaluatinglarge}, and POPE~\citep{li-etal-2023-evaluating}. These complementary tasks are outside our post-training and checkpoint-selection loop, making their aggregate score a direct check on whether visual specialization sacrifices broader capability.

\begin{table}[!t]
  \centering
  \small
  \setlength{\tabcolsep}{1.0pt}
  \renewcommand{\arraystretch}{1.12}
  \caption{Held-out generalization under benchmark-native metrics (\%, higher is better). $\operatorname{Avg}_4$ is the unweighted mean, and $\Delta$ is relative to the scale-matched Base. Within each scale, best results are bold and second-best results are underlined.}
  \label{tab:holdout-results}
  \begin{tabularx}{\columnwidth}{@{}l*{6}{>{\raggedleft\arraybackslash}X}@{}}
    \toprule
    Model & MMVP & CV & MMStar & POPE & $\operatorname{Avg}_4$ & $\Delta$ \\
    \midrule
    \rowcolor{tablegroup}
    \multicolumn{7}{c}{\itshape Qwen3.5-4B} \\
    Base & $60.00$ & $87.12$ & \bestscore{$80.27$} & $88.22$ & $78.90$ & -- \\
    GRPO & $56.00$ & $83.94$ & $69.67$ & \bestscore{$89.19$} & $74.70$ & $-4.20$ \\
    VA-OPD & $54.67$ & $85.57$ & $76.73$ & $87.77$ & $76.19$ & $-2.72$ \\
    V-Zero & \secondscore{$61.00$} & $86.70$ & $76.87$ & $88.40$ & $78.24$ & $-0.66$ \\
    Vision-OPD & $58.67$ & $86.52$ & $78.07$ & \secondscore{$88.80$} & $78.01$ & $-0.89$ \\
    Decomposed OPD & $60.67$ & \secondscore{$87.40$} & \secondscore{$79.20$} & $88.53$ & \secondscore{$78.95$} & \secondscore{$+0.05$} \\
    \rowcolor{tablebest}
    \textbf{\method{} (Ours)} & \bestscore{$62.00$} & \bestscore{$88.30$} & $77.53$ & $88.72$ & \bestscore{$79.14$} & \bestscore{$+0.24$} \\
    \midrule
    \rowcolor{tablegroup}
    \multicolumn{7}{c}{\itshape Qwen3.5-9B} \\
    Base & \secondscore{$68.67$} & $88.40$ & \bestscore{$83.80$} & $88.68$ & \secondscore{$82.39$} & -- \\
    GRPO & $66.00$ & \bestscore{$89.18$} & $81.40$ & $89.24$ & $81.46$ & \secondscore{$-0.93$} \\
    VA-OPD & $65.33$ & $83.58$ & $78.53$ & \secondscore{$89.36$} & $79.20$ & $-3.18$ \\
    V-Zero & $66.67$ & $88.18$ & $79.40$ & $87.86$ & $80.53$ & $-1.86$ \\
    Vision-OPD & $66.67$ & $84.78$ & $81.20$ & $89.10$ & $80.44$ & $-1.95$ \\
    Decomposed OPD & $67.33$ & $87.90$ & $80.10$ & $88.79$ & $81.03$ & $-1.36$ \\
    \rowcolor{tablebest}
    \textbf{\method{} (Ours)} & \bestscore{$70.00$} & \secondscore{$88.87$} & \secondscore{$82.13$} & \bestscore{$89.46$} & \bestscore{$82.62$} & \bestscore{$+0.23$} \\
    \bottomrule
  \end{tabularx}
\end{table}

\textbf{VAD preserves Base-level generalization at both scales.}
Among post-trained models, \method{} achieves the highest held-out average at both 4B ($79.14$) and 9B ($82.62$), improving over the scale-matched Base by $+0.24$ and $+0.23$ points. Decomposed OPD is the closest 4B alternative at $78.95$ ($+0.05$), but its 9B average of $81.03$ is $1.36$ points below Base; \method{} leads it by $0.19$ and $1.59$ points, respectively. \method{} also exceeds Vision-OPD by $1.13$ and $2.18$ points. Although individual benchmark leaders vary, \method{} is the only post-trained method with a positive $\Delta$ at both scales, showing that attribution-based reconstruction preserves held-out capability more consistently while improving fine-grained perception.

\subsection{RQ5: Effects of Training Choices}
\label{sec:training-choice-effects}

We examine three choices that determine how strongly and in what geometry \method{} updates the student: the weak-regularizer weight $\lambda$, the positive-branch cap $\tau_+$, and the supervision divergence.

\textbf{The training choices are robust around a broad optimum.}
The best setting, $(\lambda,\tau_+)=(0.10,0.80)$, reaches $78.32$, while nearby configurations remain within $0.20$ points; only the boundary setting $(0.00,0.50)$ drops to $76.81$. Table~\ref{tab:divergence-effects} then holds target construction and the scale-specific $(\lambda,\tau_+)$ fixed to compare symmetric JSD with the directional Forward and Reverse KL alternatives under the same official pipeline.

\begin{table}[!t]
  \centering
  \normalsize
  \setlength{\tabcolsep}{1.0pt}
  \renewcommand{\arraystretch}{1.10}
  \caption{Effect of the supervision divergence on Qwen3.5-4B and Qwen3.5-9B. Within each scale, best results are bold and second-best results are underlined.}
  \label{tab:divergence-effects}
  \begin{tabularx}{\columnwidth}{@{}l>{\columncolor{tablebest}\centering\arraybackslash}X*{2}{>{\centering\arraybackslash}X}>{\columncolor{tablebest}\centering\arraybackslash}X*{2}{>{\centering\arraybackslash}X}@{}}
    \toprule
    & \multicolumn{3}{c}{\bfseries Qwen3.5-4B} & \multicolumn{3}{c}{\bfseries Qwen3.5-9B} \\
    \cmidrule(lr){2-4}\cmidrule(lr){5-7}
    Benchmark & JSD & F-KL & R-KL & JSD & F-KL & R-KL \\
    \midrule
    V$^\star$ & $92.15$ & \bestscore{$92.72$} & \secondscore{$92.24$} & \bestscore{$94.76$} & $91.10$ & \secondscore{$93.19$} \\
    Zoom & \bestscore{$60.59$} & $59.83$ & \secondscore{$60.18$} & \bestscore{$62.49$} & \secondscore{$62.01$} & $61.42$ \\
    HR-4K & \bestscore{$85.38$} & \secondscore{$85.00$} & \secondscore{$85.00$} & \bestscore{$87.88$} & \secondscore{$87.38$} & $87.12$ \\
    HR-8K & \bestscore{$83.38$} & \secondscore{$82.75$} & $82.62$ & $85.75$ & \secondscore{$86.75$} & \bestscore{$87.00$} \\
    MME-EN & \bestscore{$76.97$} & $76.07$ & \secondscore{$76.08$} & $76.56$ & \secondscore{$76.71$} & \bestscore{$76.96$} \\
    MME-CN & \bestscore{$71.47$} & $69.32$ & \secondscore{$70.98$} & \bestscore{$72.13$} & $71.32$ & \secondscore{$71.67$} \\
    \midrule
    $\operatorname{Avg}_6$ & \bestscore{$78.32$} & $77.62$ & \secondscore{$77.85$} & \bestscore{$79.93$} & $79.21$ & \secondscore{$79.56$} \\
    \bottomrule
  \end{tabularx}
\end{table}

\textbf{JSD yields the highest aggregate accuracy at both scales.}
At 4B, JSD reaches $78.32$ $\operatorname{Avg}_6$ and leads on five benchmarks, except VStar. At 9B, it reaches $79.93$ and leads on four benchmarks; Reverse KL instead leads on HRBench-8K and MME-RealWorld. JSD is therefore the most balanced default across tasks and scales.

\FloatBarrier
\section{Conclusion and Limitations}

We introduced \method{} to address source-mixed teacher corrections in multimodal on-policy distillation. \method{} contrasts evidence-present and evidence-removed teacher views, attributes the visually aligned correction, and reconstructs a student-anchored target with separate support and refutation budgets. Across six benchmarks, it leads scale-matched methods at 4B and 9B while preserving Base-level held-out performance. Semantic and offline analyses show that the attributed direction concentrates visual and decision-relevant content while strengthening correct-token support and wrong-token suppression. This supports attribution-based reconstruction over direct source-mixed teacher distillation.

\method{} has two main limitations. First, each intervention is represented by one contrastive vector from a single view pair, which may bias compositional evidence; multiple views or learned directional bases could provide a richer estimate. Second, the current projection yields semantic enrichment rather than an identifiable separation: the attributed component can retain nonvisual teacher effects, and the residual remains source-mixed. Learned decompositions with grounding constraints may yield cleaner attribution.

\bibliographystyle{plainnat}
\bibliography{references}

\newpage
\beginappendix
\section{Visual-Alignment Diagnostic}
\label{sec:appendix-visual-alignment}

We compute the diagnostic cited in the Introduction on frozen Qwen3.5-4B and Qwen3.5-9B Base models, avoiding any confound from post-training drift. Each model generates one deterministic full-image response for the same $512$ examples using temperature $0$ and seed $42$. At every response position, we reuse that model's prefix to evaluate the full-image student, evidence-present crop teacher, and evidence-removed crop teacher. We retain the top-$128$ candidates from each view, analyze their union after renormalization, and use a missing-support log-probability floor of $-40$. This yields $64{,}340$ positions for 4B and $47{,}161$ for 9B.

On this shared support, we measure correction strength and visual alignment by
$$
D_t=D_{\mathrm{KL}}\!\left(p_T^+\Vert p_S^0\right),
\qquad
\rho_t=
\frac{\left\lVert r_t^{\mathrm{vis}}\right\rVert_2}
{\left\lVert r_t\right\rVert_2+\epsilon},
$$
where $r_t^{\mathrm{vis}}$ is the one-sided projection in Equation~\ref{eq:one-sided-attribution}. For each model $m$, we set $\tau_D^{(m)}=Q_{0.75}(\{D_t\})$ and $\tau_\rho^{(m)}=Q_{0.75}(\{\rho_t\})$, then report
$$
A^{(m)}
=
\frac{
\left|\left\{t:D_t\geq\tau_D^{(m)},\ \rho_t\geq\tau_\rho^{(m)}\right\}\right|
}{
\left|\left\{t:D_t\geq\tau_D^{(m)}\right\}\right|
}.
$$
Thus, $A^{(m)}$ is the fraction of top-quartile teacher corrections that also have top-quartile visual alignment. For 4B, $\tau_D=0.217903$ and $\tau_\rho=0.514454$, giving $A=3{,}727/16{,}085=23.2\%$. For 9B, $\tau_D=0.258831$ and $\tau_\rho=0.481173$, giving $A=2{,}687/11{,}791=22.8\%$.

\section{Controlled Post-Training Algorithms}
\label{sec:appendix-controlled-algorithms}

The controlled rows in Table~\ref{tab:main-results} share a scale-matched Qwen3.5 starting point, training set, rollout budget, and total update budget. They therefore compare where each algorithm obtains its learning signal and how that signal enters optimization, rather than differences in backbone size or data volume.

\paragraph{GRPO}
GRPO is a reinforcement-learning baseline driven by outcome reward~\citep{shao2024deepseekmathpushinglimitsmathematical}. For each prompt, it samples a group of responses, scores them with the answer-correctness reward, and normalizes these scores within the group to estimate relative advantages. A clipped policy-ratio objective then increases the likelihood of above-average responses and decreases that of below-average responses, with a reference-policy KL term controlling drift. Unlike the distillation methods below, GRPO receives trajectory-level reward rather than a privileged teacher distribution at every generated prefix.

\paragraph{Vision-OPD}
Vision-OPD transfers regional perception to a full-image policy through on-policy self-distillation~\citep{yuan2026visionopdlearningfinedetails}. The student conditions on the full image and generates the response, while a fixed teacher initialized from the same model conditions on the evidence-centered crop. Both policies evaluate each student-generated prefix, and training minimizes their token-level distributional divergence. The crop-conditioned teacher distribution is used directly as the local target, producing dense supervision without changing the target according to how much of the teacher correction is visually attributable.

\paragraph{VA-OPD}
Visual-Advantage OPD augments teacher matching with a token-level estimate of visual dependence~\citep{liu2026visualadvantageonpolicydistillationvisionlanguage}. It scores each generated token under evidence-present and evidence-degraded teacher views and rectifies their log-probability difference to obtain a nonnegative visual advantage. Trajectory-averaged advantage reweights sibling rollouts, while tokens are divided into high- and low-advantage groups whose distillation losses are normalized separately. The degraded view determines the aggregation weights, but the evidence-present teacher remains the per-position distillation target.

\paragraph{V-Zero}
V-Zero uses contrastive visual evidence to gate trajectories without answer labels~\citep{sun2026vzeroanswerlabelfreeonpolicydistillation}. For every student rollout, the teacher replays the sampled tokens with a task-relevant positive crop and an evidence-poor negative view. The mean positive-minus-negative sampled-token log-probability gap defines an evidence score, which is normalized among sibling rollouts and converted into a clipped nonnegative gate. This gate scales dense token-level distillation from the positive-view teacher: the negative view decides how strongly a rollout is learned, but it does not replace the positive teacher distribution as the local target.

\paragraph{Decomposed OPD}
Decomposed OPD factorizes multimodal distillation into language-prior and visual-grounding objectives~\citep{yoon2026decomposedonpolicydistillationvisionlanguage}. It combines the student's text-only prior with the teacher's multimodal-to-text-only log-probability shift to construct a visual target. Visual Gradient Steering then adds the resulting visual divergence to standard multimodal distillation, with gradient normalization and a language-preservation term controlling interference. Our controlled implementation retains this decomposition while matching the Qwen3.5 initialization, training examples, rollout budget, and update budget used by the other scale-matched methods.

\paragraph{\method{}}
\method{} reconstructs its target from evidence-present and evidence-removed views. At each student-generated prefix, it compares the complete privileged-teacher correction with the counterfactual response induced by visual-evidence availability, attributes the aligned component, and separates visual support from visual refutation. A budgeted asymmetric transformation then applies this attributed direction to the current student distribution, yielding a student-anchored target; a weak direct-teacher term is retained only as a stability regularizer. Thus, the visual contrast changes candidate-token target odds rather than only selecting or weighting existing teacher-matching losses.

\paragraph{Comparison boundary}
The controlled comparison isolates six distinct training signals: answer-level reward for GRPO; direct dense teacher matching for Vision-OPD; token- and rollout-level loss reweighting for VA-OPD; trajectory gating for V-Zero; language/visual target decomposition with gradient steering for Decomposed OPD; and attribution-based target reconstruction for \method{}. All implementations share the same scale-matched backbone, examples, and update budget. Section~\ref{sec:appendix-token-level-comparison} further distinguishes their local supervision mechanisms.

\section{Training Efficiency and Compute Cost}
\label{sec:appendix-training-cost}

To complement the accuracy comparison in Table~\ref{tab:main-results}, we compare training-core cost for the controlled post-training methods with comparable per-step timing logs. For each method and model scale, we use the first $65$ optimization steps, corresponding to one epoch, and report the mean step time, its standard deviation across steps, and the accumulated training time. All runs use $8$ GPUs, a global batch size of $96$, and $8$ rollouts per prompt. The training-core timer excludes benchmark evaluation, checkpoint saving, initialization, and external judge computation. We compute GPU hours as $C_{\mathrm{GPU}}=8T_{\mathrm{train}}$. Method-specific rollout context limits follow the stable configurations used to produce the evaluated checkpoints.

\begin{table}[H]
  \centering
  \small
  \setlength{\tabcolsep}{3.2pt}
  \renewcommand{\arraystretch}{1.12}
  \caption{\textbf{Training efficiency at the 4B and 9B scales for runs with comparable timing logs.} Step time is the mean $\pm$ standard deviation across $65$ optimization steps, not across random seeds. All rows use matched H800 hardware. Lower is better.}
  \label{tab:appendix-training-cost}
  \begin{tabular}{@{}lrrr@{}}
    \toprule
    Method
    & \shortstack{Avg.\ step\\time (min) $\downarrow$}
    & \shortstack{$65$-step\\time (h) $\downarrow$}
    & \shortstack{GPU\\hours $\downarrow$} \\
    \midrule
    \rowcolor{tablescaleFour}
    \multicolumn{4}{c}{\bfseries 4B Scale} \\
    GRPO & $9.34 \pm 0.59$ & $10.12$ & $81.0$ \\
    VA-OPD & $7.60 \pm 0.27$ & $8.24$ & $65.9$ \\
    V-Zero & $13.10 \pm 0.33$ & $14.19$ & $113.5$ \\
    Vision-OPD & $7.55 \pm 0.52$ & $8.18$ & $65.5$ \\
    \textbf{\method{} (Ours)} & $7.84 \pm 0.54$ & $8.49$ & $67.9$ \\
    \midrule
    \rowcolor{tablescaleNine}
    \multicolumn{4}{c}{\bfseries 9B Scale} \\
    GRPO & $12.57 \pm 0.43$ & $13.62$ & $109.0$ \\
    VA-OPD & $10.69 \pm 0.28$ & $11.59$ & $92.7$ \\
    V-Zero & $14.72 \pm 0.44$ & $15.95$ & $127.6$ \\
    Vision-OPD & $10.00 \pm 0.26$ & $10.83$ & $86.7$ \\
    \textbf{\method{} (Ours)} & $11.11 \pm 0.36$ & $12.03$ & $96.3$ \\
    \bottomrule
  \end{tabular}
\end{table}

\paragraph{Training cost and accuracy trade-off}
Across both scales, \method{} remains in the same training-cost regime as Vision-OPD and VA-OPD while achieving substantially higher fine-grained visual accuracy. At 4B, \method{} requires $7.84$ minutes per step, $8.49$ hours per epoch, and $67.9$ GPU hours---only $3.8\%$ more step time than Vision-OPD and $3.1\%$ more than VA-OPD---yet improves $\operatorname{Avg}_6$ by $2.40$ and $3.40$ points, respectively. At 9B, it requires $11.11$ minutes per step, $12.03$ hours per epoch, and $96.3$ GPU hours, corresponding to step-time increases of $11.1\%$ over Vision-OPD and $3.9\%$ over VA-OPD, while improving $\operatorname{Avg}_6$ by $3.05$ and $3.26$ points. \textbf{These results show that the accuracy gains of \method{} do not arise from a materially larger post-training budget: attributing and reconstructing the teacher correction yields a consistently stronger accuracy--compute trade-off with only modest additional cost over direct or visually weighted OPD}.

\section{Token-Level Comparison of Distillation Mechanisms}
\label{sec:appendix-token-level-comparison}

Building on the algorithm-level overview in Appendix~\ref{sec:appendix-controlled-algorithms}, this section places Vision-OPD, VA-OPD, Decomposed OPD, and \method{} in one token-level notation. The comparison concerns the supervision defined at a fixed student-generated prefix. It separates the distribution used as the target from the aggregation of losses across tokens and rollouts, since these operations need not have the same effect on the final parameter gradient.

\subsection{Unified Token Notation}

Fix a response position $t$ and a shared candidate set $V_t$. After the same tail aggregation, numerical stabilization, and normalization used in Section~\ref{sec:method}, let
$$
p_i=p_S^0(i),
\qquad
T_i=p_T^+(i),
\qquad
C_i=p_T^-(i),
\qquad i\in V_t,
$$
Here $p$, $T$, and $C$ are the full-image student, evidence-present teacher, and evidence-removed teacher distributions. We assume $p_i>0$, $T_i>0$, and $C_i>0$ on this support. Define the complete teacher-to-student log-ratio and the counterfactual visual log-ratio as
$$
a_i=\log\frac{T_i}{p_i},
\qquad
b_i=\log\frac{T_i}{C_i}.
$$
The quantity $a_i$ is the privileged teacher's prescribed log-probability correction for token $i$ relative to the student. By contrast, $b_i$ measures the fixed teacher's evidence-conditioned log-probability response for token $i$ under the same prefix, from the evidence-removed view to the evidence-present view. Centering the second quantity gives the visual-response coordinate used by \method{},
$$
u_i
=
b_i-\frac{1}{|V_t|}\sum_{k\in V_t}b_k.
$$
The common centering term has no effect on pairwise log-odds.

\subsection{Target Matching, Reweighting, and Reconstruction}

\paragraph{Vision-OPD}
Vision-OPD directly minimizes a token-level divergence between the full-image student and the evidence-present crop teacher, with the latter serving as the target~\citep{yuan2026visionopdlearningfinedetails}. Its per-position target is therefore $T$. For any $i,j\in V_t$, the induced target-to-student log-odds ratio is
$$
\log
\frac{T_i/T_j}{p_i/p_j}
=
a_i-a_j.
$$
Thus, every coordinate of the complete privileged-teacher correction can affect the local target.

\paragraph{VA-OPD}
VA-OPD evaluates the generated token $y_{n,t}$ in rollout $n$ under evidence-present and evidence-removed teacher views~\citep{liu2026visualadvantageonpolicydistillationvisionlanguage}. In the notation above, its positive visual advantage is
$$
\operatorname{VA}_{n,t}
=
\left[b_{n,t,y_{n,t}}\right]_+.
$$
The method uses this statistic at two granularities: it reweights sibling rollouts and separately normalizes the token-level losses over high- and low-advantage groups. Let $H_n$ and $L_n$ denote these groups. For the common-divergence baseline used in our comparison, write $d_{n,t}=D_{\alpha}(T_{n,t},p_{n,t})$. Using generic nonnegative group coefficients and normalized rollout weights, the aggregation has the form
$$
\mathcal{L}_{n}^{\mathrm{group}}
=
\omega_H\frac{1}{|H_n|}\sum_{t\in H_n}d_{n,t}
+
\omega_L\frac{1}{|L_n|}\sum_{t\in L_n}d_{n,t},
$$
$$
\mathcal{L}^{\mathrm{VA}}
=
\sum_n\omega_n^{\mathrm{roll}}\mathcal{L}_{n}^{\mathrm{group}},
$$
where the nonnegative weights encode the token-group and rollout-level emphasis. Each $d_{n,t}$ still uses the evidence-present teacher distribution $T_{n,t}$ as its target. VA-OPD therefore reweights the contribution of local teacher-matching terms to the aggregate update, but it does not reconstruct a different per-position target. This statement does not imply that reweighting preserves the aggregate parameter-gradient direction; token- and rollout-level weights generally redirect that gradient.

\paragraph{Decomposed OPD}
Decomposed OPD additionally evaluates the student and teacher without the image. Let $P_i^{\mathrm{text}}$ and $T_i^{\mathrm{text}}$ denote their text-only distributions, and let $T_i^{\mathrm{mm}}$ denote the teacher distribution under its multimodal condition. Its visual target is
$$
q_{T,t}^{\mathrm{DOPD}}(i)
=
\frac{
P_i^{\mathrm{text}}
\left(T_i^{\mathrm{mm}}/T_i^{\mathrm{text}}\right)
}{
\sum_{k\in V_t}
P_k^{\mathrm{text}}
\left(T_k^{\mathrm{mm}}/T_k^{\mathrm{text}}\right)
}.
$$
This target retains the student's text-only prior while injecting the teacher's multimodal information gain. The corresponding visual divergence is combined with standard multimodal distillation through Visual Gradient Steering. Decomposed OPD therefore reconstructs a target as well as steering the aggregate gradient; unlike \method{}, however, it uses a multimodal-versus-text-only contrast, does not project the privileged teacher correction onto a controlled evidence direction, and does not explicitly budget support and refutation.

\paragraph{\method{}}
\method{} uses every coordinate of the centered counterfactual response. It separates visual support from visual refutation through the branch coefficients defined in Section~\ref{sec:method}. The final coordinate shift is
$$
\delta_{t,i}^{\mathrm{VAD}}
=
\operatorname{clip}
\left(
\beta_t^+[u_i]_+
+
\beta_t^-[u_i]_-,
-c,
c
\right),
$$
where $[z]_+=\max(z,0)$ and $[z]_- = \min(z,0)$. Because common centered-log-probability offsets cancel under softmax, the reconstructed primary target has the exact token form
$$
q_{T,t}^{\mathrm{VAD}}(i)
=
\frac{
p_i\exp\left(\delta_{t,i}^{\mathrm{VAD}}\right)
}{
\sum_{k\in V_t}p_k\exp\left(\delta_{t,k}^{\mathrm{VAD}}\right)
}.
$$
Consequently,
$$
\log
\frac{
q_{T,t}^{\mathrm{VAD}}(i)/q_{T,t}^{\mathrm{VAD}}(j)
}{
p_i/p_j
}
=
\delta_{t,i}^{\mathrm{VAD}}
-
\delta_{t,j}^{\mathrm{VAD}}.
$$
The primary supervision of \method{} therefore sets target odds through the signed, budgeted counterfactual response instead of copying the complete odds of $T$. The weak direct-teacher term in the full objective remains a separate stability regularizer and is not part of $q_{T,t}^{\mathrm{VAD}}$.

\begin{table}[!b]
  \centering
  \small
  \setlength{\tabcolsep}{1.5pt}
  \renewcommand{\arraystretch}{1.16}
  \caption{\textbf{Objective-level comparison at a fixed student-generated prefix.} The per-position target describes the probability distribution inside one distillation term, not the aggregate parameter gradient or trained-policy behavior.}
  \label{tab:appendix-token-supervision-comparison}
  \begin{tabularx}{\columnwidth}{@{}l>{\raggedright\arraybackslash}p{0.21\columnwidth}X@{}}
    \toprule
    Method & Per-position target & Role of visual contrast \\
    \midrule
    Vision-OPD & $p_T^+$ & Defines the privileged target directly. \\
    VA-OPD & $p_T^+$ & Reweights rollouts and high/low-advantage token groups. \\
    Decomposed OPD & $q_{T,t}^{\mathrm{DOPD}}$ & Reconstructs a visual-information-gain target and steers the aggregate gradient. \\
    \method{} & $q_{T,t}^{\mathrm{VAD}}$ & Reconstructs signed candidate-level target odds; $p_T^+$ also appears in a separate weak anchor. \\
    \bottomrule
  \end{tabularx}
\end{table}

\subsection{Correct-Token Promotion and Wrong-Token Suppression}

For any student-anchored target constructed from a shift $\delta_i$,
$$
q_i
=
\frac{p_i\exp(\delta_i)}
{\sum_{k\in V_t}p_k\exp(\delta_k)},
$$
the absolute probability ratio is
$$
\frac{q_i}{p_i}
=
\frac{\exp(\delta_i)}
{\mathbb{E}_{k\sim p}\left[\exp(\delta_k)\right]}.
$$
Hence, the probability of token $i$ increases exactly when
$$
\delta_i
>
\log\mathbb{E}_{k\sim p}\left[\exp(\delta_k)\right],
$$
and decreases when the inequality is reversed. A negative coordinate alone is therefore insufficient to guarantee an absolute probability decrease because softmax normalization depends on all candidate coordinates.

Let $c$ be a correct answer token and $w$ an incorrect alternative. Their relative odds satisfy
$$
\log
\frac{q_c/q_w}{p_c/p_w}
=
\delta_c-\delta_w.
$$
Thus, \method{} favors the correct token over the wrong token exactly when $\delta_c>\delta_w$. Vision-OPD instead uses the complete-teacher condition $a_c>a_w$. VA-OPD retains that local target condition but reweights the position according to its rollout and token-group assignments. Because positive clipping maps a nonpositive sampled-token log-ratio to $\operatorname{VA}_{n,t}=0$, the VA score does not explicitly distinguish visual refutation from other low-advantage cases. The low-advantage group still receives distillation loss, so this observation is a limitation of the score's attribution semantics rather than a claim that VA-OPD cannot learn suppression.

\subsection{Scope of the Comparison}

The identities above are algebraic statements about supervision at fixed prefixes. They do not by themselves establish causal differences between trained checkpoints. Figure~\ref{fig:target-effects} isolates the immediate correct-token support and wrong-token suppression of the effective supervision directions; VA-OPD has no separate direction because its local target remains the same $p_T^+$ target used by Vision-OPD, while its weighting acts during loss aggregation and optimization. Table~\ref{tab:main-results} compares the complete trained systems under a common evaluation pipeline. Accordingly, the frozen-state analysis characterizes local objective behavior, whereas the trained-policy table measures the accumulated optimization outcome.

\end{document}